# Attack Assessment and Augmented Identity Recognition for Human Skeleton Data

Joseph G. Zalameda
*Dept. of ECE*
Old Dominion University
Norfolk, VA, USA
jzala001@odu.edu

Megan A. Witherow
*Dept. of ECE*
Old Dominion University
Norfolk, VA, USA
mwith010@odu.edu

Alexander M. Glandon
*Dept. of ECE*
Old Dominion University
Norfolk, VA, USA
aglan001@odu.edu

Jose Aguilera
*Dept. of Computer Science*
Amherst College
Hampshire County, MA, USA
jaguilera23@amherst.edu

Khan M. Iftekharuddin
*Dept. of ECE*
Old Dominion University
Norfolk, VA, USA
kiftekha@odu.edu

*Abstract*—Machine learning models trained on small data sets for security applications are especially vulnerable to adversarial attacks. Person identification from LiDAR based skeleton data requires time consuming and expensive data acquisition for each subject identity. Recently, Assessment and Augmented Identity Recognition for Skeletons (AAIRS) has been used to train Hierarchical Co-occurrence Networks for Person Identification (HCN-ID) with small LiDAR based skeleton data sets. However, AAIRS does not evaluate robustness of HCN-ID to adversarial attacks or inoculate the model to defend against such attacks. Popular perturbation-based approaches to generating adversarial attacks are constrained to targeted perturbations added to real training samples, which is not ideal for inoculating models with small training sets. Thus, we propose Attack-AAIRS, a novel addition to the AAIRS framework. Attack-AAIRS leverages a small real data set and a generative adversarial network (GAN) generated synthetic data set to assess and improve model robustness against unseen adversarial attacks. Rather than being constrained to perturbations of limited real training samples, the GAN learns the distribution of adversarial attack samples that exploit weaknesses in HCN-ID. Attack samples drawn from this distribution augment training for inoculation of the HCN-ID to improve robustness. Ten-fold cross validation of Attack-AAIRS yields noticeable increase in robustness to unseen attacks- including Fast Gradient Sign Method, Projected Gradient Descent, Additive Gaussian Noise, Momentum Iterative Fast Gradient Sign Method, and Basic Iterative Method. The HCN-ID Synthetic Data Quality Score for Attack-AAIRS indicates that generated attack samples are of similar quality to the original benign synthetic samples generated by AAIRS. Furthermore, inoculated models show consistent final test accuracy with the original model trained on real data, demonstrating that our method improves robustness to adversarial attacks without reducing testing performance on real data.

## I. INTRODUCTION

Efficient 3D skeleton encoding of human form and pose have proven effective for human behavioral analysis and identification with applications in security and surveillance [1-5]. In particular, far field LiDAR based skeleton extraction systems [1] enable long range skeleton acquisition in low light environments as opposed to visual based skeleton extraction [6]. However, LiDAR based skeleton data sets are expensive and time-consuming to collect, and thus, suffer from a limited number of samples, noise in joint movements between video frames, error in joint location estimation (e.g., occluded joint locations), and null frames. Overcoming these challenges is paramount to ensure stringent reliability and robustness of security systems.

Deep learning methods such as Convolutional Neural Networks (CNNs) have shown robust performance on skeleton data for security-related tasks, including action recognition [2,5,7], gender classification [4,8,9], and person identification [1,3,9]. Hierarchical Co-occurrence Networks (HCN) [5], which leverage the local- and global-level feature aggregation properties of CNN, have proven effective for action recognition [5] and person identification tasks (HCN-ID) [3]. However, like other deep learning approaches, HCN is prone to overfit to small training data and requires large data sets to accurately approximate the true decision boundary [3]. In [3], Generative Adversarial Network (GAN) [10] based data augmentation, e.g., Deep Convolution GAN

(DC-GAN) [11] and Auxiliary Classifier GAN (AC-GAN) [12], are shown to improve HCN-ID performance on relatively small person identification data sets based on LiDAR skeleton extraction.

Recently, adversarial attack methods threaten models deployed in security applications and vulnerable systems where model reliability and robustness must be well understood. Adversarial attacks may take place at training time, e.g., 'poisoning' the training set, or test time [13]. We focus on test time, where attacks seek to exploit weaknesses in the model's learned decision boundaries [14]. Such attacks may be especially potent in the case of small data sets, with which models may suffer poorly regularized decision boundaries due to insufficient training data.

A popular mode of test time attack is the generation of structured noise, i.e., a perturbation, that is added to the input to cause misclassification. One of the earliest and most popular examples of this approach is the Fast Gradient Sign Method (FGSM) [14], which uses the model's gradients to generate an attack perturbation to shift samples across the decision boundary and cause misclassifications. Other examples of perturbation-based attack methods include Projected Gradient Descent (PGD) [15], Gaussian Noise (GN) [16], Momentum Iterative Fast Gradient Sign Method (MI-FGSM) [17], and Basic Iterative Method (BIM) [18]. A popular defense framework is to train models on perturbed input samples to inoculate the model against known attacks [13]. Therefore, to generate more varied and complex perturbations, GAN-based approaches, such as Unrestricted Black-box Adversarial Attack Using GAN [19] and Attack Inspired GAN (AI-GAN) [20], have become increasingly popular. However, perturbation-generating methods [14-19] are constrained by the available input samples to which the attack noise will be added. Thus, such methods suffer from the challenges associated with learning from small data, e.g., overfitting. Furthermore, generation of new attack methods is an active research area and thus, it may not be possible to inoculate models against all possible attacks. Recently, Adversarial Transformation GAN (AT-GAN) [21] has been proposed to learn a distribution of adversarial examples close to the real data distribution, yielding synthetic adversarial examples that are not constrained to perturbations of real input examples. The real data distribution is learned using an AC-GAN [12] and then transferred to attack a target model [21].

Synthetic adversarial data generation and inoculation has not been studied for person identification in small LiDAR extracted skeleton data sets. Existing attack frameworks for skeleton data, such as the Constrained Iterative Attack for Skeleton Actions (CIASA) [22], rely on constrained adversarial examples generated by perturbing real skeleton data and fail to address challenges associated with small data sets. Given that adversarial attacks exploit weaknesses in a model's decision boundaries which are further exacerbated in models trained on small data sets [14], we anticipate that AC-GAN based data augmentation as in [3] will enrich the decision boundaries by better capturing the underlying real data distribution. Furthermore, we hypothesize that the distribution of adversarial samples learned by an AT-GAN trained to exploit a model's weaknesses may be used to fortify decision boundaries against a wide variety of attacks.

Inspired by the success of AC-GAN for synthetic skeleton data augmentation [3], we propose an AT-GAN based attack generation and inoculation framework to yield a more robust HCN-ID based person identification model. We evaluate the performance of inoculated and un-inoculated models on multiple unseen attack methods, including FGSM, PGD, GN, MI-FGSM, and BIM. Building on the existing Assessment of Augmented Identity Recognition for Skeletons (AAIRS) framework from [3], we name our proposed method Attack Assessment of Augmented Identity Recognition for Skeletons (Attack-AAIRS). The contributions of Attack-AAIRS are as follows:

- The vulnerability of a target classifier (HCN-ID) trained on a small LiDAR extracted skeleton data set is assessed considering multiple attack methods (FGSM, PGD, GN, MI-FGSM, and BIM).
- An AT-GAN is trained to produce synthetic adversarial examples that are not constrained to the small real LiDAR extracted skeleton data set. The quality of these synthetic adversarial examples is assessed in comparison to synthetic samples generated from the data distribution learned from real data.
- The synthetic adversarial examples are used to inoculate the target classifier. The inoculated model is evaluated considering multiple unseen attack methods (FGSM, PGD, GN, MI-FGSM, and BIM).
- Inoculated and un-inoculated classifiers are evaluated on real test data to study the effect of inoculation on model classification performance.

Attack-AAIRS provides a reliable framework for vulnerability assessment and attack inoculation of skeleton person identification models given a small skeleton data set extracted from LiDAR videos.

## II. BACKGROUND

### A. Skeleton Data

Consider a series of skeleton data frames as a skeleton video. Each frame is represented by a 3-dimensional point cloud. Each point is a spatial location of a joint at a given moment in time. Each skeleton video sample takes the form of a tensor $X$, which has a 4-dimensional structure (Eq. 1) with N samples, C=3 axes, T frames, and J joints. Each sample has a label denoting the associated person ID. The shape of the label tensor $Y$ is shown in Eq. 2.

$$Size(X) = [N, C, T, J] \quad (1)$$

$$Size(Y) = [N, 1] \quad (2)$$

*B. HCN-ID*

HCN-ID [3] is an HCN [5] network trained to identify people in skeleton videos based on their gait. HCN [5] is a CNN-based deep network for classification of skeleton videos based on spatial movement of individual joints through time. HCN $H(x_1, x_2, ..., x_N|\theta_H)$ takes as input the model weights $\theta_H$ and a skeleton video sample with multiple frames $x_1, x_2, ..., x_N \in X \in \mathbb{R}^{C,T,J}$. A HCN models joint co-occurrence by performing global feature aggregation across the skeleton joints as channels during convolution [5]. To explicitly model temporal motion of the skeleton, two parallel CNN branches process (1) the current joint locations and (2) the difference between the joint locations in the current frame and the previous frame, respectively [5]. Eq. 3 describes how the temporal difference in the latter branch is computed [5]:

$$\dot{x} = x(:,1:T,:) - x(:,0:T-1,:) \quad (3)$$

This results in a derivative computation that is filtered along with the original branch. The outputs of these two branches are concatenated and then further processed by a set of CNN and feedforward classification layers.

*C. AAIRS Framework*

The AAIRS framework [3] has been applied to assess viability of synthetic data augmentation on a small skeleton dataset extracted from multiple LiDAR videos exhibiting noise in joint movement between video frames, error in joint location estimation, and null frames. The framework [3] leverages AC-GAN [12] to augment training of HCN-ID. Training augmentation involves learning the underlying real data distribution using AC-GAN [12]. Then, the trained generator from AC-GAN samples this distribution to produce synthetic samples that are mixed in with the real samples to augment the model training. As reported in [3], AAIRS-based data augmentation achieves an increase in testing accuracy of the HCN-ID of 5.68% with a mix of 50% real and 50% synthetic data across 10 cross validation folds.

Quality of the produced synthetic skeletons is quantified in [3] using an adaptation of inception score called the HCN-ID score. The HCN-ID score range is [1,9], with 1 being the lowest and 9 being the highest [3]. The synthetic skeletons produced by the AAIRS framework exhibit an average score of 3.933 ± 0.867. The real data yields a mean HCN-ID score of 8.352 ± 0.105.

To our knowledge, neither HCN-ID or the AAIRS framework have been assessed for adversarial attack vulnerability.

*D. Attack Frameworks*

FGSM [14] is popular test time attack method that alters inputs with a small, targeted perturbation. A successful attack will cause the targeted classifier to misclassify the perturbed input. The perturbation is computed using the sign of the gradients of the target classifier with respect to the input image. The adversarial attack samples $X_{Attack}$ are generated by the FGSM in [14] by adding perturbation $P$ to unaltered input data $X$ with an attack magnitude value $\varepsilon$:

$$X_{FGSM} = X + \varepsilon * P \quad (4)$$

The attack methods PGD [15], BIM [18], and MI-FGSM [17] utilize a variant of the iterative based FGSM based attack framework shown in Eq. 5 where *itr* refers to the number of iterations. The Additive Gaussian Noise based attack (GN) [16] adds Gaussian noise of a fixed magnitude to the real samples.

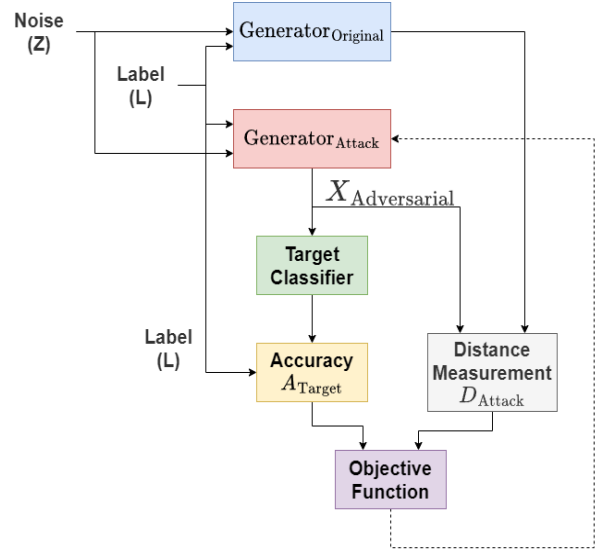

Fig. 1. AT-GAN Training Pipeline from [21]

$$x_{Attack}^{itr+1} = x_{Attack}^{itr} + \varepsilon * P \quad (5)$$

*E. AT-GAN*

The Adversarial Transformation GAN (AT-GAN) from [21] is a transfer learning framework designed to shift a pretrained generator from producing reinforcing synthetic samples to producing adversarial synthetic examples. This is done by introducing a domain shift to the pretrained generator that is designed to minimize the distance between synthetic samples produced by the original generator and adversarial synthetic samples produced by the attack generator along with a minimization in target classifier accuracy. This results in a generator that can produce unconstrained adversarial examples [21].

The AT-GAN framework consists of three individual networks: an original generator ($Generator_{Original}$), an attack generator ($Generator_{Attack}$), and a target classifier. The target classifier serves as a representation of the model to be attacked, the attack generator is the network to be trained by the AT-GAN to produce the synthetic adversarial examples, and the original generator is used as a reference of benign samples during training. The parameters for $Generator_{Original}$ and $Generator_{Attack}$ are initialized using a single source generator pretrained on the original real dataset. The target classifier is pretrained to yield high classification accuracy for both the original real dataset and the synthetic data produced by the source generator.

Once pretraining is completed, the source generator parameter values are used to initialize the original and attack generators of the AT-GAN. Feedback from the pretrained target classifier is used during the training of the AT-GAN in order to exploit that classifier's weaknesses. The trainable parameters of the original generator and target classifier are frozen for the duration of the AT-GAN training. The following objective function is used to train AT-GAN [21]:

$$ATGAN\ Objective = \text{Min}(\alpha * A_{Target} + \beta * D_{Attack}) \quad (6)$$

The goal of $Generator_{Attack}$ is to produce the synthetic adversarial examples $X_{Adversarial}$ that are designed to attack the target classifier. The objective function of the AT-GAN shown in Eq. 6 minimizes the target classifier accuracy on these adversarial samples $A_{Target}$ and the distance $D_{Attack}$ between the outputs of $Generator_{Original}$ and $Generator_{Attack}$ to ensure the generated adversarial samples are close to the original data distribution. The values of $\alpha$ and $\beta$ are hyperparameters designed to weight each component of the objective function.

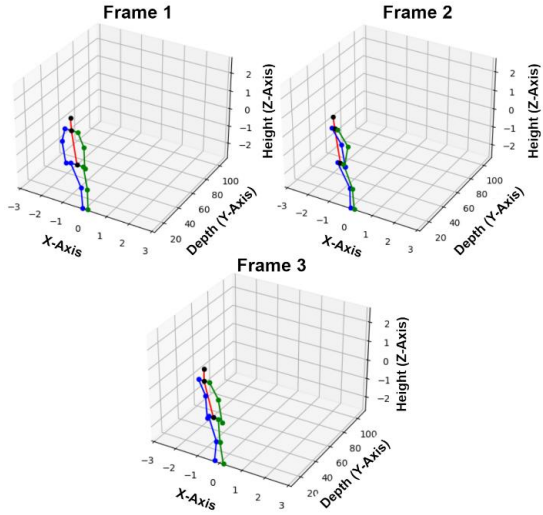

Fig. 2. Single Skeleton Gait Representation 3 Frame Sample

## III. METHODS

### A. Skeleton Data Set

We use a skeleton data set extracted from LiDAR videos of 9 people walking at various ranges [1]. A 3-frame skeleton gait representation of the connected joints along with spatial dimension labels is shown in Fig 2. Each skeleton has J=13 joints [1]. Following [3], we sample the videos into T=3 frame segments. The resulting class distribution is shown in Fig 3.

Additionally, each subject is centered at the origin and rotated to always face the front of the scene. All joint locations are normalized between [-1,1].

### B. Attack-AAIRS Framework

The proposed Attack-AAIRS framework builds on the existing AAIRS framework [3] to (1) assess attack vulnerability of the HCN-ID network to multiple attack frameworks, (2) learn the distribution of adversarial attack samples for HCN-ID using the AT-GAN approach, generate and assess the quality of adversarial attack samples drawn from this distribution, (3) inoculate HCN-ID for test time attacks by training the network with generated adversarial examples and assess the robustness of the inoculated HCN-ID to multiple unseen attack methods (FGSM, PGD, GN, MI-FGSM, and BIM), (4) study the effect of inoculation on HCN-ID model accuracy.

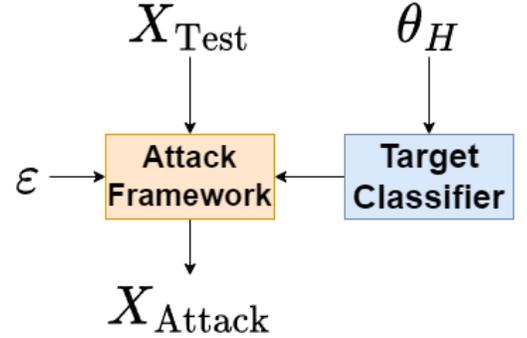

Fig. 4. Attack-AAIRS Attack Test Data Pipeline

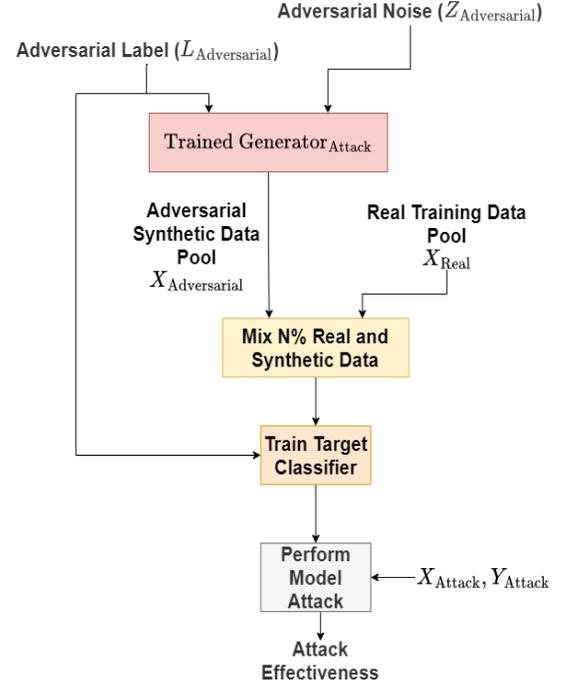

Fig. 5. Attack-AAIRS Augmentation Pipeline

The four goals above are implemented as follows. For (1), we create an attack test set $X_{Attack}$ to assess vulnerability to a given attack method (Fig 4). We apply the attack method with perturbation magnitude value $\varepsilon$ to real test samples $X_{Test}$ with test set labels $Y_{test}$. We consider an untargeted attack, where attack set labels $Y_{Attack}$ are the same as real test set labels $Y_{Test}$.

We perform in depth analysis on the FGSM [14] because it is one of the most common attack methods and the basis for many others (e.g., PGD, MI-FGSM, and BIM). We also evaluate three iterative methods (PGD, MI-FGSM, and BIM) as well as simple additive gaussian noise (GN).

The quality of the resulting $X_{Attack}$ samples are evaluated as the attack success rate, measured as the percentage of attack samples misclassified by the target classifier. For (2), we use the AT-GAN framework (Fig 1) to yield a $Generator_{Attack}$ capable of producing unconstrained synthetic adversarial examples $X_{Adversarial}$ that exploit the vulnerabilities of HCN-ID. We consider the pretrained generator from the AAIRS framework as $Generator_{Original}$. $Generator_{Attack}$ is initialized with the weights of $Generator_{Original}$. Training minimizes Eq. 6 to produce adversarial examples $X_{Adversarial}$ from the input adversarial noise $Z_{Adversarial}$ and adversarial labels $L_{Adversarial}$. To quantitatively assess the quality of the generated samples, we use the HCN-ID Score [3], a quality score for synthetic skeleton samples based on inception score [23]. For (3), we perform inoculation (Fig 5) by augmenting the training of HCN-ID with samples from $Generator_{Attack}$. The robustness of HCN-ID to each unseen attack method (FGSM, PGD, GN, MI-FGSM, or BIM) before and after inoculation is performed using $(X_{Attack}, Y_{Attack})$. Accuracy on real test samples $(X_{Test}, Y_{Test})$ is also assessed for (4).

*C. Experiments*

The proposed Attack-AAIRS framework requires a pretrained target classifier and pretrained AC-GAN generator $Generator_{Original}$. For the target classifier, we consider two variants of the HCN-ID from the AAIRS framework: (a) HCN-ID Model 1 trained on only real skeleton data and (b) HCN-ID Model 2 trained on a N=50% mix of real data and synthetic data from the AAIRS generator. These HCN-ID variants represent the baseline model and the best model, respectively, reported in [3]. The process for pretraining these two models is shown in Fig 6. Note that the left most branch of Fig 6 follows the same structure as the AAIRS framework [3]. For both variants, we train following [3] using a batch size of 64 for 100 epochs. The gradients are updated using the Adam optimizer [24] with a learning rate of 0.001. For $Generator_{Original}$, we consider the AAIRS generator trained following [3] using a batch size of 32 for 600 epochs. The $Generator_{Original}$ network structure is extended with an additional convolutional layer in addition to the network in [3]. We train $Generator_{Attack}$ for 100 epochs with a batch size of 900 evenly distributed across all 9 classes. The $Generator_{Attack}$ parameters are updated using Stochastic Gradient Descent (SGD) with a learning rate of 0.0002. The ideal values for $\alpha$ and $\beta$ are empirically determined as 1 and 2, respectively. We assess the proposed Attack-AAIRS framework using multiple unseen attack methods (FGSM, PGD, GN, MI-FGSM, or BIM). Using FGSM as an exemplar ($X_{Attack} = X_{FGSM}$), we provide further in-depth assessment of Attack-AAIRS robustness. A good attack method will be effective (in terms of attack success rate) and produce attack samples that appear benign. Thus, we assess and compare the quality of Attack-AAIRS generated samples to benign samples generated by AAIRS. A good defense method is expected to improve robustness to attacks without degrading performance on real samples. Therefore, we also compare the performance of inoculated and un-inoculated models on real data.

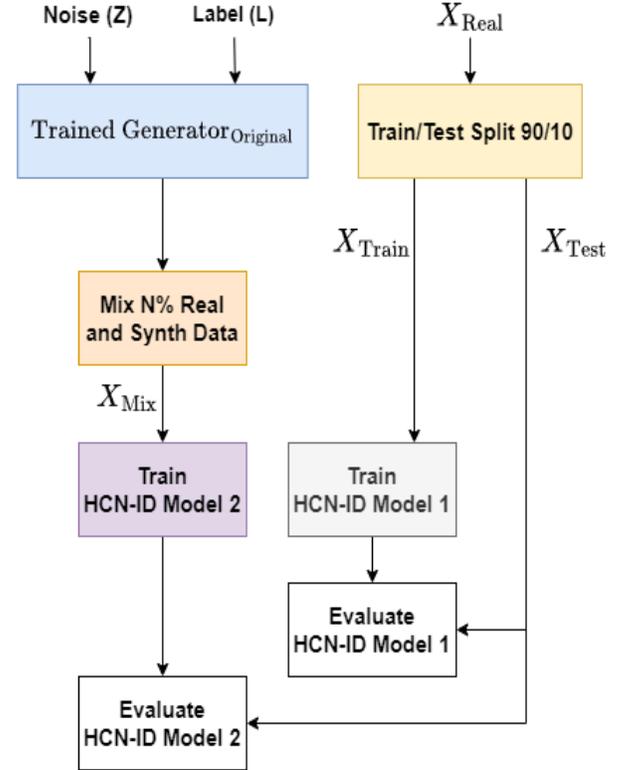

Fig. 6. HCN-ID Target Classifier Pretraining

We consider several experiments as follows:

1. We benchmark the vulnerability of the two variants of the HCN-ID for multiple attack frameworks. For FGSM, we study attack success rate across different perturbation magnitudes $\varepsilon$. Multiple attack frameworks are used in Fig 4 to generate $X_{Attack}$. We repeat this procedure for the inoculated models to assess the effect of augmented training with samples from $Generator_{Attack}$. We quantify model robustness as $1 - Attack\ Success\ Rate$.

2. The HCN-ID score [3] is used to assess the quality of synthetic samples from $Generator_{Attack}$ compared to $Generator_{Original}$ and the real data.

3. Accuracy of the inoculated and un-inoculated versions of HCN-ID Model 1 and HCN-ID Model 2 are evaluated on the real test data.

We repeat all experiments over 10 cross validation folds where each fold represents 10% of the data set and all classes are represented. Each fold is held out for testing once, yielding ten different 90%/10% train/test splits. To prevent cross contamination between the train and test sets, models are randomly initialized between each fold.

## IV. RESULTS AND DISCUSSION

### A. Attack-AAIRS Results

We benchmark HCN-ID Model 1 and HCN-ID Model 2

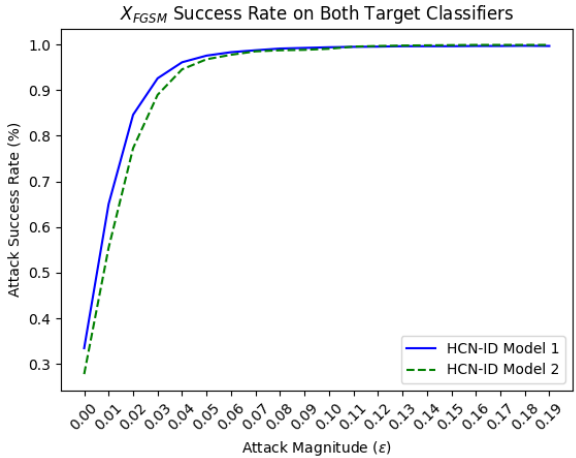

Fig. 7. Attack-AAIRS Effectiveness on Pretrained HCN-ID Models using FGSM

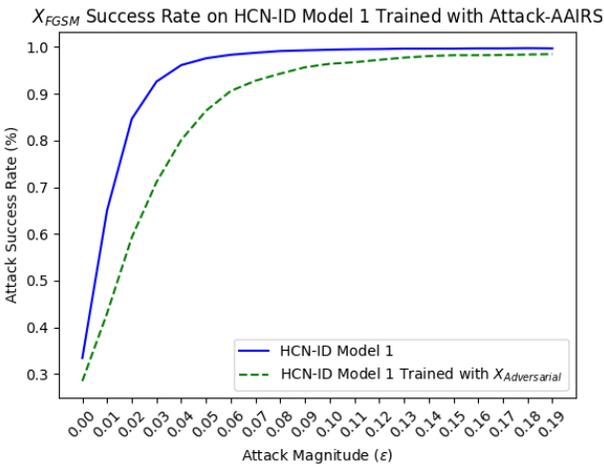

Fig. 8. Attack-AAIRS Augmentation Pipeline Effectiveness on HCN-ID Model 1 using FGSM

against $X_{Attack}$ to measure inherent robustness prior to inoculation. Fig 7 plots the mean FGSM attack success rate over 10 folds for different attack magnitudes $\varepsilon$ ranging from 0 to 0.19. The initial attack magnitude $\varepsilon$=0 corresponds to the baseline test performance of the model without any perturbation to the test set. HCN-ID Model 2 exhibits better baseline robustness to the attack compared to HCN-ID Model

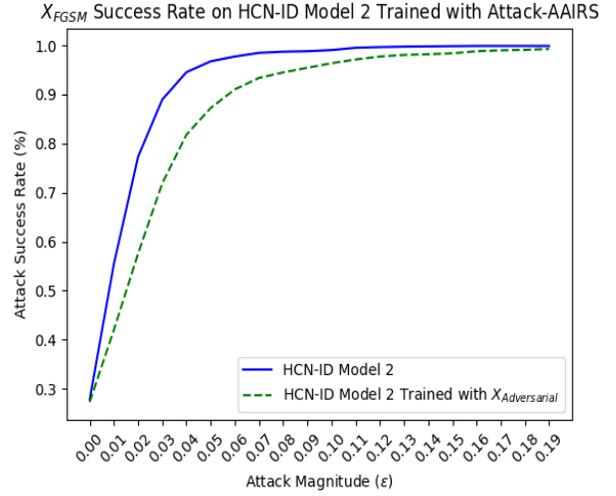

Fig. 9. Attack-AAIRS Augmentation Pipeline Effectiveness on HCN-ID Model 2 using FGSM

1.

We attribute this increased robustness to the AC-GAN based synthetic data training augmentation in the AAIRS framework [3]. Due to the limited number of samples in the LiDAR skeleton data set, it is expected that HCN-ID Model 1 does not have access to the full real data distribution during training, yielding a less accurate and more vulnerable decision boundary.

The synthetic samples generated by the AC-GAN are expected to fill in gaps in the real data training distribution, resulting in improved test performance and robustness to adversarial attacks for HCN-ID Model 2. The attack effectiveness converges at $\varepsilon = \sim 0.1$ to be roughly equal for both models. This indicates room for improvement against more aggressive attacks for both HCN-ID Model 1 and 2.

Fig 8 and Fig 9 compare FGSM based $X_{Attack}$ effectiveness on HCN-ID Model 1 and HCN-ID Model 2, respectively, before and after inoculation with the adversarial samples from $Generator_{Attack}$. A noticeable improvement in robustness can be seen for the inoculated models over all $\varepsilon$ values. The most notable improvement for both inoculated models is shown when $\varepsilon \leq 0.1$. Given that HCN-ID Model 2 is more resilient to $X_{Attack}$ prior to inoculation (Fig 7), the magnitude of improvement is not as substantial for HCN-ID Model 2 (Fig 9) compared to HCN-ID Model 1 (Fig 8).

Table I. Attack-AAIRS Average Model Robustness Improvement

| Attack Method | Robustness Improvement on HCN-ID Model 1 | Robustness Improvement on HCN-ID Model 2 |
|---|---|---|
| FGSM | 0.071 | 0.054 |
| PGD | 0.083 | 0.061 |
| GN | 0.240 | 0.010 |
| MI-FGSM | 0.067 | 0.050 |
| BIM | 0.135 | 0.110 |

Table 1 shows the mean improvement in model robustness ($1 - Attack\ Success\ Rate$) to $X_{Attack}$ for both HCN-ID Model 1 and HCN-ID Model 2 for multiple attack frameworks. The associated parameters for each framework are as follows: PGD [15] with $eps = \frac{8}{255}, alpha = \frac{2}{255}, steps = 10$; GN with $magnitude = 1, stdev = 0.1$; MI-FGSM with $eps = \frac{8}{255}, alpha = \frac{2}{255}, steps = 10, decay = 1.0$; and BIM [18] with $eps = \frac{8}{255}, alpha = \frac{2}{255}, steps = 10$. This shows that the model inoculation with Attack-AAIRS provides better robustness for both target classifiers across multiple attack frameworks.

### B. Quality Assessment of Adversarial Samples

An important property of effective attack methods is that adversarial attack samples appear similar to benign real samples. Thus, we assess the quality of samples produced by Attack-AAIRS $Generator_{Attack}$ using the HCN-ID Score [3] and compare with the scores from $Generator_{Original}$ and the real data. These results are reported in Table 2. The HCN-ID Score is computed using HCN-ID Model 1, which is trained only on real skeleton data. HCN-ID Score ranges from [1,9] with a score of 1 being the worst and a score of 9 being the best.

Note that the improvements to the network structure of $Generator_{Original}$ yield an increase in the mean HCN-ID Score of 0.764 over the reported HCN-ID Score from [3]. Furthermore, although the focus of the Attack-AAIRS is adversarial training, the final HCN-ID Score produced by the Attack-AAIRS generated data only degrades the HCN-ID score by 0.244. This indicates that the adversarial transformation learned by the Attack-AAIRS framework maintains skeleton quality close to the original synthetic samples.

Table II: Mean HCN-ID Score Assessment

| Data Set | Mean HCN-ID Score |
|---|---|
| Real Data (Baseline) | $8.381 \pm 0.067$ |
| Synthetic Data from AAIRS [3] | $3.933 \pm 0.867$ |
| Synthetic Data from Attack-AAIRS original generator | $4.348 \pm 0.880$ |
| **Synthetic Data from Attack-AAIRS adversarial generator** | **$4.104 \pm 0.799$** |

### C. Evaluation on Real Test Data

A good defense framework will improve a model's robustness to attacks and also maintain model performance on the real data. The mean test accuracy over 10 cross validation folds for the real test data ($X_{Test}, Y_{Test}$) is reported in Table 3. The worst performing model is HCN-ID Model 1 trained with only real data. The inoculated HCN-ID Model 2 is the best performing model, achieving a test accuracy of 72.74% which is a slight accuracy increase over HCN-ID Model 1. This is achieved by training with both the AAIRS and Attack-AAIRS frameworks. The inclusion of the Attack-AAIRS framework not only improves model robustness to $X_{Attack}$ but also maintains test accuracy on the real data. Attack-AAIRS targets weaknesses in the HCN-ID decision boundaries and patches these weaknesses through the inoculation process. This yields a more robust decision boundary that is less vulnerable and better captures the real underlying data distribution.

Table III. Mean Testing Accuracy Across all 10 Folds

| Network | Mean Test Accuracy |
|---|---|
| **Base HCN-ID** (HCN-ID Model 1) | $66.54\% \pm 3.29\%$ |
| **HCN-ID Trained with Attack-AAIRS** (HCN-ID Model 1 Trained with $X_{Adversarial}$) | $71.47\% \pm 3.58\%$ |
| **HCN-ID Trained with AAIRS** (HCN-ID Model 2) | $72.22\% \pm 3.29\%$ |
| **HCN-ID Trained with AAIRS+Attack-AAIRS** (HCN-ID Model 2 Trained with $X_{Adversarial}$) | **$72.74\% \pm 3.70\%$** |

## V. CONCLUSION AND FUTURE WORK

The proposed Attack-AAIRS provides an adversarial attack training and testing framework for use with small, gait based human identity classification skeleton data sets extracted from LiDAR videos. This method leverages the GAN-based synthetic data generation to augment adversarial training of the target classifier, which allows for meaningful classifier robustness improvements appropriate for small data sets. Furthermore, these improvements are achieved while maintaining synthetic skeleton data quality. Combining the AAIRS and Attack-AAIRS frameworks for training improves model accuracy and robustness to adversarial attacks, both of which are critical for security applications.

In the future, the Attack-AAIRS framework will be used to benchmark model robustness to more complex adversarial attacks [13] such as GAN-based attacks [21, 22]. The modular design of Attack-AAIRS is designed to make interchanging attack frameworks simple and allow for relatively quick adaptation and evaluation of different attack types. Among these, we plan to study the robustness of models trained with Attack-AAIRS against AT-GAN generated adversarial test sets learned using independent sets of LiDAR extracted skeleton data. We also plan to study the feasibility of using improved AC-GAN based architectures like the Fast-Converging Conditional Generative Adversarial Network [25] along with this Attack-AAIRS framework. Additionally, we plan to comprehensively study the properties of decision boundaries in small data problems and the effect of GAN-based methods on improving the quality of such boundaries along with comparing with other existing state of the art adversarial defense methods [26].

ACKNOWLEDGEMENTS

The authors would like to acknowledge partial support of this work by US Army NVESD, CERDEC through Grant No. 100659, DoD Center of Excellence in AI and Machine Learning (CoE-AIML) under Contract Number W911NF-20-2-0277 with the U.S. Army Research Laboratory, and by the National Science Foundation under Grant No. 1828593, Grant No. 1950704, and Grant No. 1753793.